\documentclass[letterpaper, 10 pt, conference]{ieeeconf}  

\IEEEoverridecommandlockouts                              

\usepackage{gensymb}
\usepackage[dvipsnames]{xcolor}
\usepackage{makecell}
\usepackage{multirow}

	\newcommand{\V}[1]{ {\boldsymbol{#1}}}  

	\newcommand{\MAT}[1] {\begin{bmatrix} #1 \end{bmatrix}}
	
 \usepackage{cancel}
 \usepackage[normalem]{ulem}

\usepackage{array}
\newcolumntype{Y}{>{\centering\arraybackslash}X}

\usepackage{graphicx} 
\usepackage{amsmath} 
\newtheorem{theorem}{Theorem}
\usepackage{amssymb}  
\usepackage{breqn}
\usepackage{tabularx}
\usepackage{caption}
\usepackage{subcaption}
\usepackage{booktabs} 
\usepackage[export]{adjustbox}
\usepackage{subcaption}
\usepackage{xcolor}
\usepackage{hyperref}
\usepackage{cleveref}
\usepackage{cite}

\crefformat{equation}{#2eq.~#1#3}
\Crefformat{equation}{#2Eq.~#1#3}

\crefmultiformat{equation}{#2eqs.~#1#3}{ and~#2#1#3}{, #2#1#3}{ and~#2#1#3}
\Crefmultiformat{equation}{#2Eqs.~#1#3}{ and~#2#1#3}{, #2#1#3}{ and~#2#1#3}

\crefrangeformat{equation}{#3eqs.~#1#4 to~#5#2#6}
\Crefrangeformat{equation}{#3Eqs.~#1#4 to~#5#2#6}

\title{\LARGE \bf
Robotic Valve Turning: Axial Misalignment Correction Using Reaction Torque Feedback}

\author{Amit Kumar$^{1}$, Sri Harsha Turlapati$^{2}$, Gautami Golani$^{1}$, Yang Lin$^{1}$, Ravi N. Banavar$^{3}$, and Domenico Campolo$^{4}$ 
\thanks{This research is supported by the National Research Foundation, Singapore, under the NRF Medium Sized Centre scheme (CARTIN). Any opinions, findings, and conclusions or recommendations expressed in this material are those of the authors and do not reflect the views of National Research Foundation, Singapore.}
\thanks{$^{1}$ Amit Kumar, Gautami Golani and Yang Lin are with the School of Mechanical and Aerospace Engineering, Nanyang Technological University, Singapore.
        }%
\thanks{$^{2}$Sri Harsha Turlapati is with the School of Science and Technology, Singapore University of Social Sciences, Singapore.
        {\tt\small sriharsha@suss.edu.sg}}%
\thanks{$^{3}$Ravi N. Banavar is with the Centre for Systems and Control, Indian Institute of Technology Bombay, Mumbai, India.
        {\tt\small banavar@iitb.ac.in}}%
\thanks{$^{4}$Domenico Campolos is with the R3 Institute (R3I), and School of Mechanical and Aerospace Engineering, Nanyang Technological University, Singapore.
        {\tt\small d.campolo@ntu.edu.sg}}%
}

\begin{document}

\maketitle
\thispagestyle{empty}
\pagestyle{empty}

\begin{abstract}

In this work, we propose a haptic update control law that uses reaction torques to correct axial misalignment during robotic valve manipulation. Unlike vision-based estimates, which can be affected by calibration errors, occlusion, and uncertainty in the contact geometry, reaction torques arise directly from the physical interaction between the gripper and valve. A geometric relationship exists between the error (misalignment) vector and these torques. The primary aim of this work is to propose a stable controller exploiting this geometric property. Our control law is proven to be uniformly asymptotically stable. Simulations are performed for verification. Furthermore, we experimentally test the robustness of our method using a Kinova Gen3 robotic arm for initial misalignments ranging from $-15\degree$ to $15\degree$ at 3 different valve positions and report the resulting data distribution. The absolute value of the median misalignment across all 18 test cases is found to be within $2.46\degree$ and that of reaction torques within $0.23\mathrm{Nm}$.

\end{abstract}

\section{INTRODUCTION}

Accurate valve turning is a benchmark manipulation task because it is a common requirement across many industries, such as oil \& gas, and nuclear power plants~\cite{2013_nuclear_jap,DARPA_why_valve}. It is even common in many household tasks, such as turning the knob of a gas stove or adjusting the thermostat; an important task for the humanoid butlers~\cite{Nikos_2014, Schaffer_2026}.

Ajoudani et al. in~\cite{Nikos_2014} developed a compliant humanoid robot (COMAN) for accurate valve turning. Their algorithm updates the Cartesian stiffness profiles of the two arm endpoints based on friction along the valve axis to avoid high interaction forces. Brunner et al.~\cite{Tognon_2022} have attached a rigid end-effector to an aerial vehicle (drone) to turn the valve. They developed a cost function comprising the pose, object, distance to the object, and force from an augmented system state, and used the Model Predictive Path Integral Control strategy. Faria et al.~\cite{Faria_2015} used three vision cameras, one each mounted on an arm of a dual-arm robot and one at the head, to manipulate a single valve. The valve has point-light markers attached. For 20 different valve positions, they reported a maximum orientation estimation error of approximately $5.72\degree$ (0.1 rad). Camera calibration errors and the reliance on fiducial markers can introduce pose estimation errors in such setups, motivating the use of wrench feedback for accurate alignment. Few studies have investigated using reaction forces and torques for valve turning~\cite{2024_react_force}. Xing et al.~\cite{2024_react_force} have proposed a passivity-based force tracking method for axial displacement correction using a wheeled mobile manipulator. They created a compliant end-effector that utilizes full measured interaction wrench (both forces and torques) to correct the misalignment. In contrast, our proposed method achieves axial misalignment correction by relying exclusively on reaction torques.

\begin{figure}[tb]
    \centering
    \includegraphics[width=0.85\linewidth]{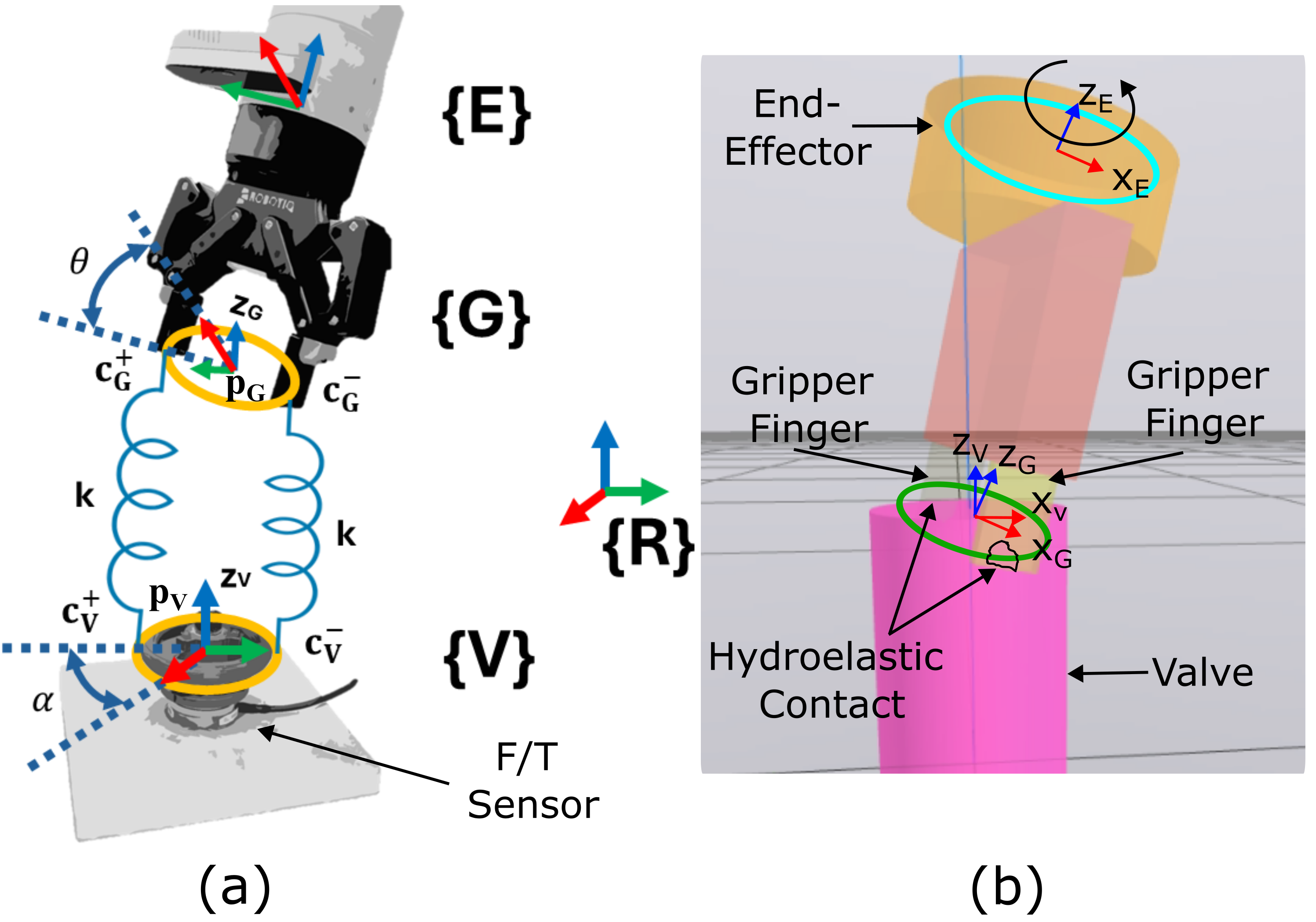}
    \caption{(a) Valve interaction model showcasing the misalignment between valve $(z_V)$ and gripper $(z_G)$ axes. (b)~Drake simulation setup: Hydroelastic contact between the fingers and the valve generates reaction torques used for online pose correction.}
    \label{fig:valve_model_v2}
\end{figure}

Humans' or animals' ability to manipulate physical objects vastly exceeds that of modern robots, especially for contact-rich tasks~\cite{Campolo_2025,hogan_2023}. It has been shown that humans use impedance (inverse of compliance) control to achieve high dexterity~\cite{2001_Kawato}. Moreover, impedance superimposes linearly; a useful property for redundant manipulators~\cite{2022_Hogan}. Reproducing such adaptability in automated systems remains challenging. Compliance is therefore often introduced, for example, using flexible couplers, to accommodate unavoidable kinematic discrepancies. Building on such notions, a quasi-static model to predict reaction torques as a function of the axis misalignment is presented in~\cite{Gautami2024} and shown in Figure~\ref{fig:valve_model_v2}(a). Linear springs, with stiffness coefficients $k$, are used to model the interaction force between the gripper's fingers and the valve's contact points. Golani et al.~\cite{Gautami2024} concluded that the misalignment vector is tangent to the reaction torques ellipses and its length is proportional to the magnitude of the reaction torques. 

This work focuses on a haptic update control law that reduces misalignment using these geometric relationships. The system is shown to be persistently exciting and uniformly asymptotically stable. It is then tested in the Drake simulator~\cite{drake}, which is well suited to contact-rich tasks. Finally, we test the robustness of our control law using a 7-DoF Kinova Gen3 robotic arm across 18 test cases with varying initial misalignments ranging from $-15\degree$ to $15\degree$ at three different valve positions. The rest of the paper is structured as follows: Section~\ref{sec:metho} describes the methodology of our valve-interaction model, haptic update control law, and the robotic arm controller, Section~\ref{sec:stability_analysis} proves the stability of our controller, Section~\ref{sec:drake_simulation} details the drake simulation along with its results, and Section~\ref{sec:hw_exp} presents the hardware experiment with its results. The notations used in the paper are as follows:

The rotation matrix, $\mathbf{R} \in SO(3)$, is defined as:
\begin{align*}
    SO(3) &= \{ \mathbf{R} \in \mathbb{R}^{3 \times 3} : \mathbf{R}\mathbf{R}^{\top} = \mathbf{I}, \det \mathbf{R} = +1 \}
\end{align*}
and its associated Lie algebra is defined as:
\begin{align*}
    \mathfrak{so(3)} &= \{ \mathbf{R} \in \mathbb{R}^{3 \times 3} : \mathbf{R} + \mathbf{R}^{\top} = \mathbf{0}, \operatorname{tr}(\mathbf{R}) = 0 \}
\end{align*}
Let \( \boldsymbol{\Omega} \in \mathbb{R}^3 \), then we define $\hat{\boldsymbol{\Omega}}$ as,
\begin{align*}
    \hat{\boldsymbol{\Omega}} =
    \begin{pmatrix}
    0 & -\Omega_3 & \Omega_2 \\
    \Omega_3 & 0 & -\Omega_1 \\
    -\Omega_2 & \Omega_1 & 0
\end{pmatrix} \in \mathfrak{so(3)}
\end{align*}
For any \( v \in \mathbb{R}^3 \) then \( \hat{\boldsymbol{\Omega}} \boldsymbol{v} = \boldsymbol{\Omega} \times \boldsymbol{v} \) is the vector cross product. Similarly, the operator $(.)^\vee$: $\mathfrak{so(3)} \to \mathbb{R}^3$ denotes the inverse of the hat map.

\section{Methodology}
\label{sec:metho}

\subsection{Valve Interaction Model~\cite{Gautami2024}}
\label{sec:valve_interaction_model}
We assume the valve frame $\{V\}$ is at the origin, with $\MAT{\V x_V & \V y_V & \V z_V}$ denoting unit vectors and $\V p_V = \MAT{0 & 0 & 0}^\top$. The gripper-frame misalignment is parameterized with respect to the valve's vertical axis using $n_1$ and $n_2$ as:
\begin{equation}
    \V z_G = \frac{1}{\sqrt{n_1^2+n_2^2+1}}\MAT{n_1\\n_2\\1}, \V p_G = \MAT{d_1\\d_2\\d_3}
    \label{eq:zg_def}
\end{equation}
The gripper frame is further defined by $\V x_G$ and $\V y_G$ as:
\begin{equation}
    \V y_G = \frac{\V z_G \times \V x_V}{||\V z_G \times \V x_V||}, \V x_G = \V y_G \times \V z_G
    \label{eq:yg_xg_def}
\end{equation}
with the frame rotation given by: 
\begin{align}
    \V R_G = \MAT{\V x_G & \V y_G & \V z_G} \in SO(3)
    \label{eq:R_G}
\end{align}
and the frame location at $\V p_G$. The contact points on the gripper are parameterized by $\theta$ and gripper radius $r_G$:
\begin{align}
    \V c_G^+(\theta) &= r_G(\cos(\theta)\V x_G +\sin(\theta)\V y_G) + \V p_G,\label{eq:cg_p_def} \\
    \V c_G^-(\theta) &= \V c_G^+(\theta + \pi)
    \label{eq:cg_m_def}
\end{align}
Similarly, we have the valve contact points:
\begin{align}
    \V c_V^+(\alpha) &= r_V(\cos(\alpha)\V x_V +\sin(\alpha)\V y_V) + \V p_V,\label{eq:cv_p_def}\\
    \V c_V^-(\alpha) &= \V c_V^+(\alpha + \pi)
    \label{eq:cv_m_def}
\end{align}
The strain energy due to contact modeled via elastic deformation of linear springs (along the direction of $\V z_G$) with stiffness $k$ between respective contact pairs is:
\begin{equation}
    E(\alpha,\theta) = \frac{k}{2}\left(||\V c_G^+(\theta)-\V c_V^+(\alpha)||^2+||\V c_G^-(\theta)-\V c_V^-(\alpha)||^2\right)
    \label{eq:strain_e_def}
\end{equation}
For a freely rotating valve (i.e., without friction and inertia), the quasi-static condition yields:
\begin{align}
    \nabla_\alpha E &= 0
    \label{eq:quasi_static_strain_energy}
\end{align}
For a given gripper rotation command $\theta$, we solve eq. (9) to compute the valve rotation $\alpha^*$ under equilibrium conditions:
\begin{align}
    \alpha^*(\theta,n_1,n_2) = \tan^{-1}\frac{\sqrt{n_1^2+n_2^2+1}\sin(\theta)-n_1n_2\cos(\theta)}{(1+n_2^2)\cos(\theta)}
    \label{eq:alpha_star}
\end{align}
We compute the force acting on the gripper contacts:
\begin{align}
    \V f^+(n_1,n_2,\theta) &= k(\V c_G^+(\theta) - \V c_V^+(\alpha^*)),\label{eq:f_plus}\\
    \V f^-(n_1,n_2,\theta) &= k(\V c_G^-(\theta) - \V c_V^-(\alpha^*)) \label{eq:f_minus}
\end{align}
and the torque acting on the gripper:
\begin{equation}
    \V \tau_{G}(n_1,n_2,\theta) = \V c_G^+ \times \V f^+ + \V c_G^- \times \V f^-
    \label{eq:tau_G_n1_n2_theta}
\end{equation}

\subsection{Haptic Update Control Law}
\label{sec:haptic_update_rule}
In this setup, the gripper rotates at a constant angular velocity, $\dot{\theta}=\omega^G$, while maintaining its grasp on the valve. As reaction torques scale with axial misalignment, the evolution of the gripper axis $\V z_G$ is governed by a first-order differential equation designed to continuously minimize these torques. The ODE is defined as:
\begin{equation}
    \dot{\V z}_G = - \beta \cdot \V \tau_G(n_1,n_2,\theta) \times \V z_G.
    \label{eq:z_G_dot}
\end{equation}
where $\beta$ is the scalar rotational admittance with units~$\mathrm{(Nm}~\cdot\mathrm{s})^{-1}$. Differentiating eq.\ref{eq:yg_xg_def} we get:
\begin{align}
    \dot{\mathbf{y}}_{G} &= \frac{\dot{\tilde{\mathbf{y}}}_{G} - \mathbf{y}_{G} (\mathbf{y}_{G} \cdot \dot{\tilde{\mathbf{y}}}_{G})}{\|\mathbf{z}_{G} \times \mathbf{x}_V\|}, \label{eq:y_G_dot} \\
    \dot{\mathbf{x}}_{G} &= (\dot{\mathbf{y}}_{G} \times \mathbf{z}_{G}) + (\mathbf{y}_{G} \times \dot{\mathbf{z}}_{G}) \label{eq:x_G_dot}
\end{align}
where $\dot{\tilde{\mathbf{y}}}_{G}= (\dot{\mathbf{z}}_{G} \times \mathbf{x}_V) + (\mathbf{z}_{G} \times \dot{\mathbf{x}}_V)$ and $\dot{\mathbf{x}}_V = \mathbf{0}$. The rate of change of frame rotation is then:
\begin{align}
    \dot{\V R}_G = \MAT{\dot{\V x}_G & \dot{\V y}_G & \dot{\V z}_G}
    \label{eq:R_G_dot}
\end{align}

\subsection{Robotic Arm Control}
\label{sec:robotic_arm_control}
Consider the robotic arm to have $n$ degrees of freedom. Its base frame is denoted by $\{R\}$. The rigid body angular velocity of $\{G\}$ in $\{R\}$ is computed using~\cref{eq:R_G,eq:R_G_dot} as:
\begin{align}
    \hat{\mathbf{\Omega}} &= \dot{\mathbf{R}}_G \mathbf{R}_G^{\top} \in \mathfrak{so}(3) \label{eq:Omega} \\
    \boldsymbol{\omega}^R &= (\mathbf{R}_R^V)^{\top} (\hat{\mathbf{\Omega}})^{\vee} \in \mathbb{R}^{3} 
    \label{eq:omega_in_R}
\end{align}
where, $\mathbf{R}_R^V$ is the rotation matrix from frame $\{R\}$ to $\{V\}$. The required angular velocity is mapped to the joint velocity via the inverse of the Jacobian. The velocity of first ~$(n~-~1)$ joints are updated using the inverse of the reduced geometric Jacobian ($J_{\text{reduced}} \in \mathbb{R}^{(n-1) \times (n-1)}$).  The $n^{th}$ joint is updated independently based on its specified spin velocity $\omega^G$.

\begin{align}
\dot{\mathbf{q}}^{(1:(n-1))} &= J_{\text{reduced}}^{-1} \begin{bmatrix} \mathbf{0}_{3 \times 1} \\ \boldsymbol{\omega}^R \end{bmatrix} \nonumber \\
\dot{q}^{(n)} &= \omega^G
\label{eq:q_t_update}
\end{align}

\section{Stability Analysis}
\label{sec:stability_analysis}
In terms of free variables, $n_1$ and $n_2$, the system in~\cref{eq:z_G_dot} is given by~\cref{eq:app_n1_dot,eq:app_n2_dot} from Appendix~\ref{sec:app_haptic_up_rule_n_dot}. Our system is hence of the form:
\begin{align}
    \dot{\V n} = f(n, u(t))
    \label{eq:n_dot_dyna}
\end{align}
where $u(t)=\theta = \omega^G t$ is the input of the system. Its equilibrium is given by:
\begin{align}
    &\dot{\V n}= 0 
    \label{eq:dot_n_equilibrium}\\
    \implies &\V \tau_G =0 , \V z_G =0, \ \text{or } (\V \tau_G \times \V z_G) =0 \nonumber
\end{align}
Note that, $\V z_G \neq 0$ because of our definition in~\cref{eq:zg_def}. When $\V \tau_G$ is parallel to $\V z_G$, $(\V \tau_G \times \V z_G) =0$. However, this is also not possible because $\V \tau_G$ is perpendicular to $\V z_G$ in our model. Using~\cref{eq:app_n1_dot,eq:app_n2_dot}, the equilibrium of the system is: 
\begin{align}
    \bar{\V n} = (n_1,n_2) = (0, 0)
    \label{eq:equilibrium_n_dot}
\end{align}
Note that $\tan\left(\theta \right) = \frac{-n_1}{n_2\,\sqrt{n_1^2 +n_2^2 +1}}$ is not an equilibrium even though this condition leads to $\dot{\V n}= 0$. Equilibrium point is defined as a point where the system states can stay forever~\cite{slotine1991applied}. This condition will only hold at a particular time instant. However, this condition causes our system to be persistently excited, as shown in Section~\ref{sec:drake_simulation}. To prove the stability of our system, we use the following theorem from~\cite{Annaswamy_1987}:

\begin{theorem}~\cite{Annaswamy_1987}
\label{theo:persistent_systems}
The equilibrium state of $\dot{x} = f(x,t)$ is uniformly asymptotically stable in the large if a function $V(x, t)$ defined for all $x$ and $t$ with $V(0, t) = 0$ is (i) positive-definite, (ii) decrescent, (iii) radially unbounded, and (iv) $\dot{V}(x, t)$ evaluated along the trajectories of $\dot{x} = f(x,t)$ is negative-semidefinite and (v)
    $\int_{t}^{t+T} \dot{V}(x(\tau), \tau) \, d\tau \le -\gamma(\|x(t)\|) < 0$
for some $T$ and all $t \ge t_0$, where $\gamma(\cdot)$ is a positive monotonic function with $\gamma(0) = 0$.
\end{theorem}

We will now show that all the required conditions (i-v) are satisfied for the system given in eq.~\ref{eq:n_dot_dyna}. 

\noindent(i-iii) We define the candidate Lyapunov function:
\begin{align}
    V (\V n) = \frac{1}{2}\V n^\top \V n
    \label{eq:lyapunov_fn}
\end{align}
Clearly, $V(\bar{\V n}) =0$ and $V(\V n) > 0$ in $D - \{\bar{\V n}\}$, where $D = \{(n_1, n_2) \in \mathbb{R}^2\}$ making it positive definite. As $\lVert n \rVert \to \infty$, $V(\V n) \to \infty$. Hence $V(\V n)$ is radially unbounded. Also, it is decrescent as $V(\V n)$ does not depend on $t$~\cite{slotine1991applied}.

\noindent (iv) The derivative of the Lyapunov function:
\begin{align}
    \dot{V}(\V n, t) = \V n^\top\dot{\V n}
    \label{eq:dot_v}
\end{align}

\noindent Using~\cref{eq:subs_dot_v} from Appendix~\ref{sec:deri_lyapunov_func}, $\dot{V}(\V n,t) \le 0$, i.e., it is negative semi-definite as $(\rho_1, \rho_2, \rho_3, \rho_4^2)~\ge~0 \ \forall \ (n_1, n_2, \theta)~\in~\mathbb{R}^3$.

\noindent(v) Again, using~\cref{eq:subs_dot_v} from Appendix~\ref{sec:deri_lyapunov_func}, we get:
\begin{align}
    \int_t^{t+T} \dot{V}(\tau) d\tau &= -\int_t^{t+T} \Gamma(\V n, \tau) \cdot(\mathbf{w}^{\top}(\V n, \tau)\V n)^2 d\tau \\
    &= -\int_t^{t+\frac{2\pi}{\omega^G}} \Gamma(\V n, \tau) \cdot \Big(n_1\cos(\omega^G \tau)+ \nonumber \\
    & \quad \quad n_2\sqrt{n_1^2+n_2^2+1}\sin(\omega^G \tau)\Big)^2 d\tau    
    \label{eq:cond_5_persistent}
\end{align}

\begin{figure}[tb]
    \centering
    \includegraphics[width=0.85\linewidth]{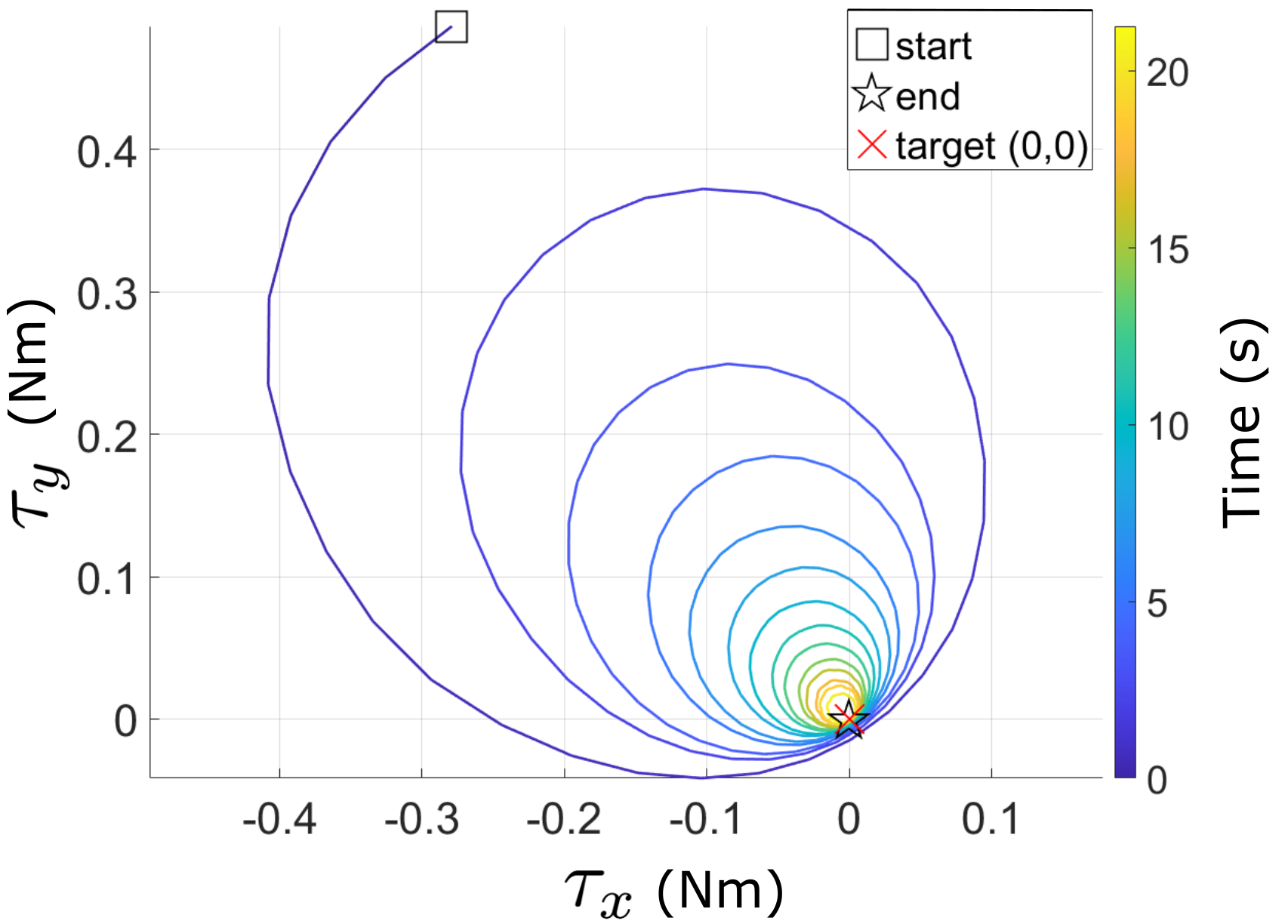}
\caption{Evolution of $\tau_x$ vs $\tau_y$ in Drake. The reaction torque ellipses contract toward the origin.}
\label{fig:drake_representative}
\end{figure}

\noindent As $\Gamma(\V n, t) = \frac{2\beta k r_G r_V \rho_3}{\rho_1 \rho_2}$, we can find its minimum with respect to $\theta$ by finding the maximum of $\rho_2$. Note that $\rho_1$ and $\rho_3$ are independent of $\theta$. It is straightforward to verify that $\Gamma(\V n, t) \ge \frac{\sqrt{\rho_3}}{\rho_{1}^2}$. Assuming that the state dynamics are sufficiently slow such that $\mathbf{n}(\tau) \approx \mathbf{n}(t) \  \forall \ \tau \in [t, t+T] $ over the short integration window $T = \frac{2\pi}{\omega^G}$:

\begin{align}
    \int_t^{t+T} \dot{V}(\tau) d\tau & \le -\frac{\sqrt{\rho_3}}{\rho_{1}^2} \int_t^{t+\frac{2\pi}{\omega^G}} \Big(n_1\cos(\omega^G \tau)+ \nonumber \\
    & \quad \quad n_2\sqrt{n_1^2+n_2^2+1}\sin(\omega^G \tau)\Big)^2 d\tau \nonumber \\
    \implies \int_t^{t+T} \dot{V}(\tau) d\tau & \le \frac{-\pi}{\omega^G} \frac{\sqrt{n_1^2 + n_2^2 + 1}}{n_2^2 + 1} \nonumber \\ 
    & \qquad \times (n_1^2+n_2^2)(n_2^2+1) \nonumber \\
    \implies \int_t^{t+T} \dot{V}(\tau) d\tau & \le - \sqrt{n_1^2 + n_2^2 + 1} \frac{\pi (n_1^2+n_2^2)}{\omega^G} \nonumber \\   
    & \le -\gamma(\vert{}\vert{}\V n(t)\vert{}\vert{})
    \label{eq:cond_5_persistent_1}
\end{align}
where, $\gamma(\vert{}\vert{}\V n(t)\vert{}\vert{})=\frac{\pi}{\omega^G} \vert{}\vert{}\V n(t)\vert{}\vert{}^2$. Clearly, $\gamma(\cdot)$ is monotonically increasing, $\gamma(0) = 0$, $\gamma(\vert{}\vert{}\V n\vert{}\vert{}) > 0 \  \forall \ \V n \ne \V 0$.

\section{Simulation using Drake}
\label{sec:drake_simulation}
A Drake~\cite{drake} simulation was developed to evaluate the proposed axial misalignment correction method, stated in~\cref{eq:z_G_dot,eq:y_G_dot,eq:x_G_dot,eq:R_G_dot,eq:Omega,eq:omega_in_R,eq:q_t_update}, under contact-rich valve turning conditions. The simulated system consists of a two-finger gripper, a valve body, and hydroelastic contact between the fingers and the valve surface, as shown in Figure~\ref{fig:valve_model_v2}. The valve frame is fixed, while the gripper is initialized with a prescribed axial misalignment parameterized by $n_1$ and $n_2$. During the simulation, the gripper continuously rotates about its own axis, and the reaction wrench generated by contact is recorded.
\begin{figure}[tb]
    \centering
    \includegraphics[width=0.85\linewidth]{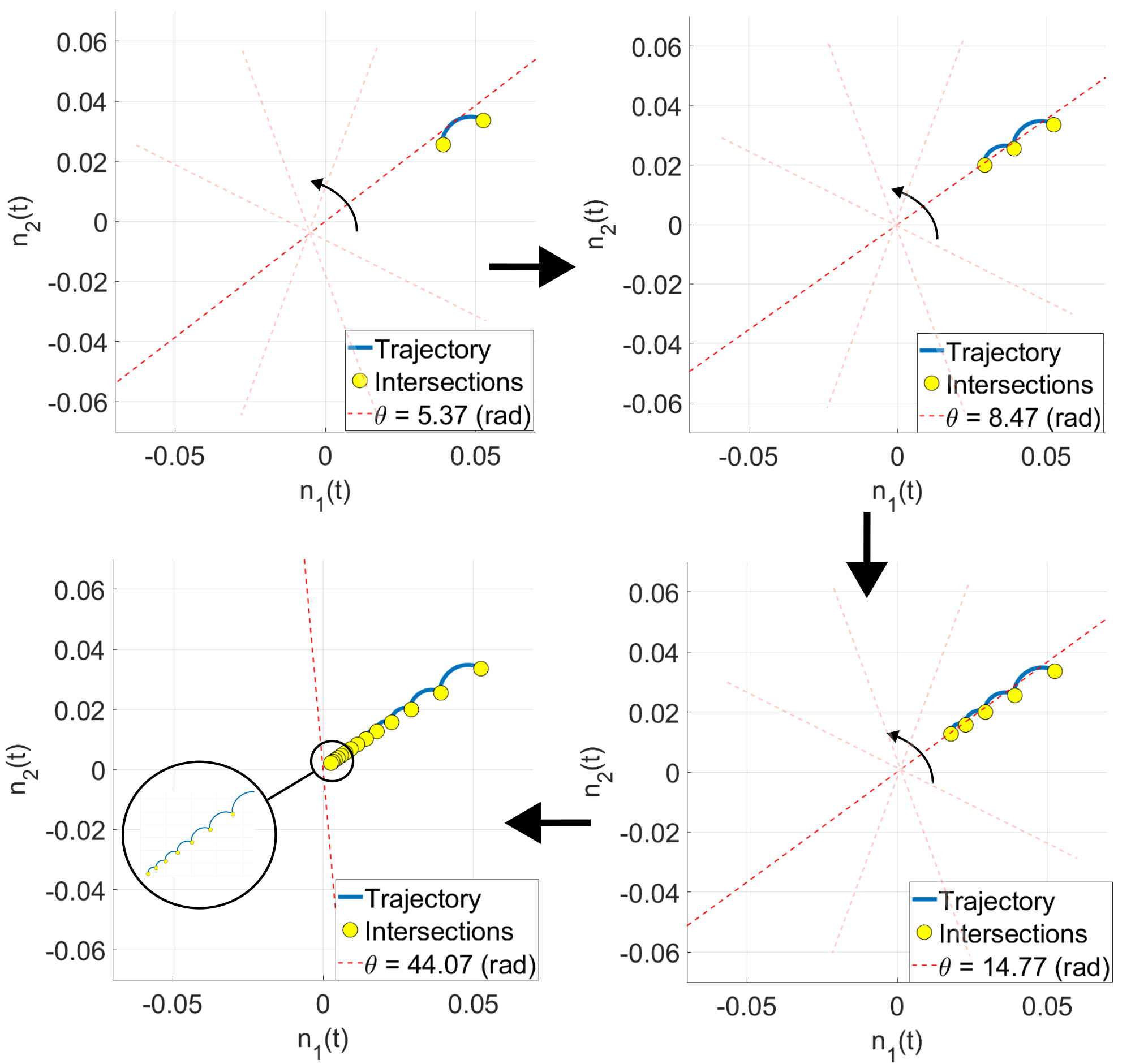}
    \caption{Evolution of $n_1$ vs $n_2$ during the axial misalignment correction in Drake. The dashed red line represents the implicit plot of the function $\tan\left(\theta \right) - \frac{-n_1}{n_2\,\sqrt{n_1^2 +n_2^2 +1}} =0$ and the yellow blobs represent the intersection of the implicit function and the trajectory of $n_1$ and $n_2$.}
    \label{fig:drake_lyapunov_n1_n2_intersec}
\end{figure}
\begin{figure}[tb]
    \centering
    \includegraphics[width=0.85\linewidth]{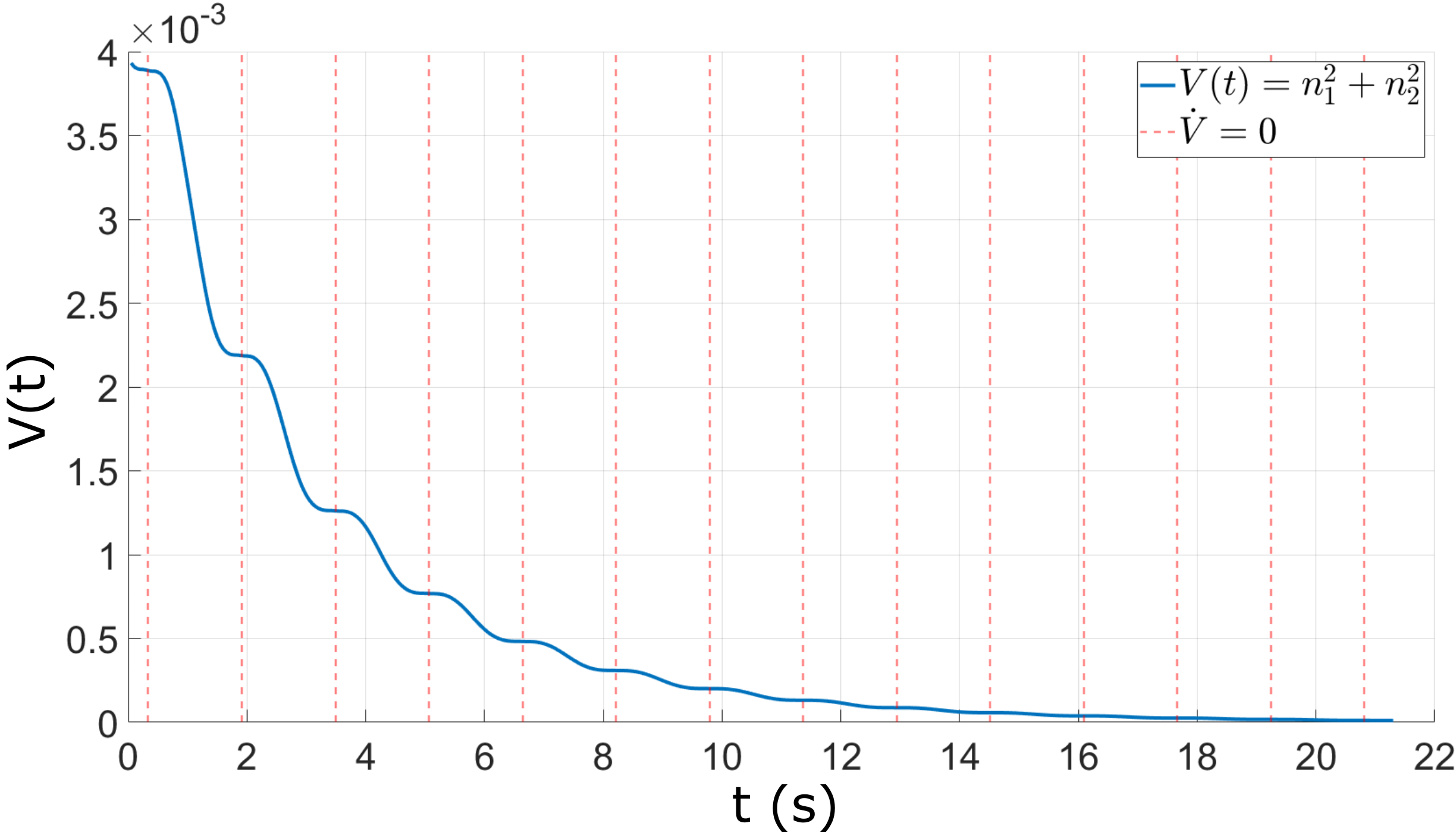}
    \caption{(Blue) Evolution of $V(t)$ with respect to time. (Red) Time instants when $\dot{V}(t) = 0$ in Drake simulation.}
    \label{fig:drake_lyapunov}
\end{figure}

At each feedback step, the gripper pose is updated based on the current values of $n_1$ and $n_2$, and a quasi-static simulation is executed to obtain the contact reaction wrench after the system reaches equilibrium. The wrench is then transformed into the valve frame, and only the transverse torque components, $\tau_x$ and $\tau_y$, are used for the system update using~\cref{eq:z_G_dot,eq:y_G_dot,eq:x_G_dot,eq:R_G_dot,eq:Omega,eq:omega_in_R,eq:q_t_update}.

Figure~\ref{fig:drake_representative} shows the evolution of $\tau_x$ vs $\tau_y$ in the Drake simulation. The torque plane trajectory initially forms a large ellipse, indicating a strong transverse reaction torque caused by misalignment. As the feedback update progresses, the ellipse shrinks toward the origin. This shows that the contact-induced lateral torque decreases over successive rotations. The corresponding evolution of $n_1$ and $n_2$ is shown in Figure~\ref{fig:drake_lyapunov_n1_n2_intersec}.

Figure~\ref{fig:drake_lyapunov_n1_n2_intersec} shows the persistently exciting nature of our system. The kinks appear repeatedly as the gripper rotates ($\theta = \omega^G t$), axial misalignment is corrected as per~\cref{eq:z_G_dot,eq:y_G_dot,eq:x_G_dot,eq:R_G_dot,eq:Omega,eq:omega_in_R,eq:q_t_update}, and the system evolves from Figure~\ref{fig:drake_lyapunov_n1_n2_intersec}(a-d). The dashed red line represents the implicit plot of the function $\tan\left(\theta \right) - \frac{-n_1}{n_2\,\sqrt{n_1^2 +n_2^2 +1}} =0$, which represents the condition when $\dot{V}=0$. The intersection of the implicit function and the state trajectory is denoted by a yellow blob. Each time it intersects, a kink is produced. The system eventually approaches close to $(0,0)$. The evolution of $V(t)$ with respect to time is shown in Figure~\ref{fig:drake_lyapunov}.

\begin{figure}[tb]
    \centering
    \includegraphics[width=0.85\linewidth]{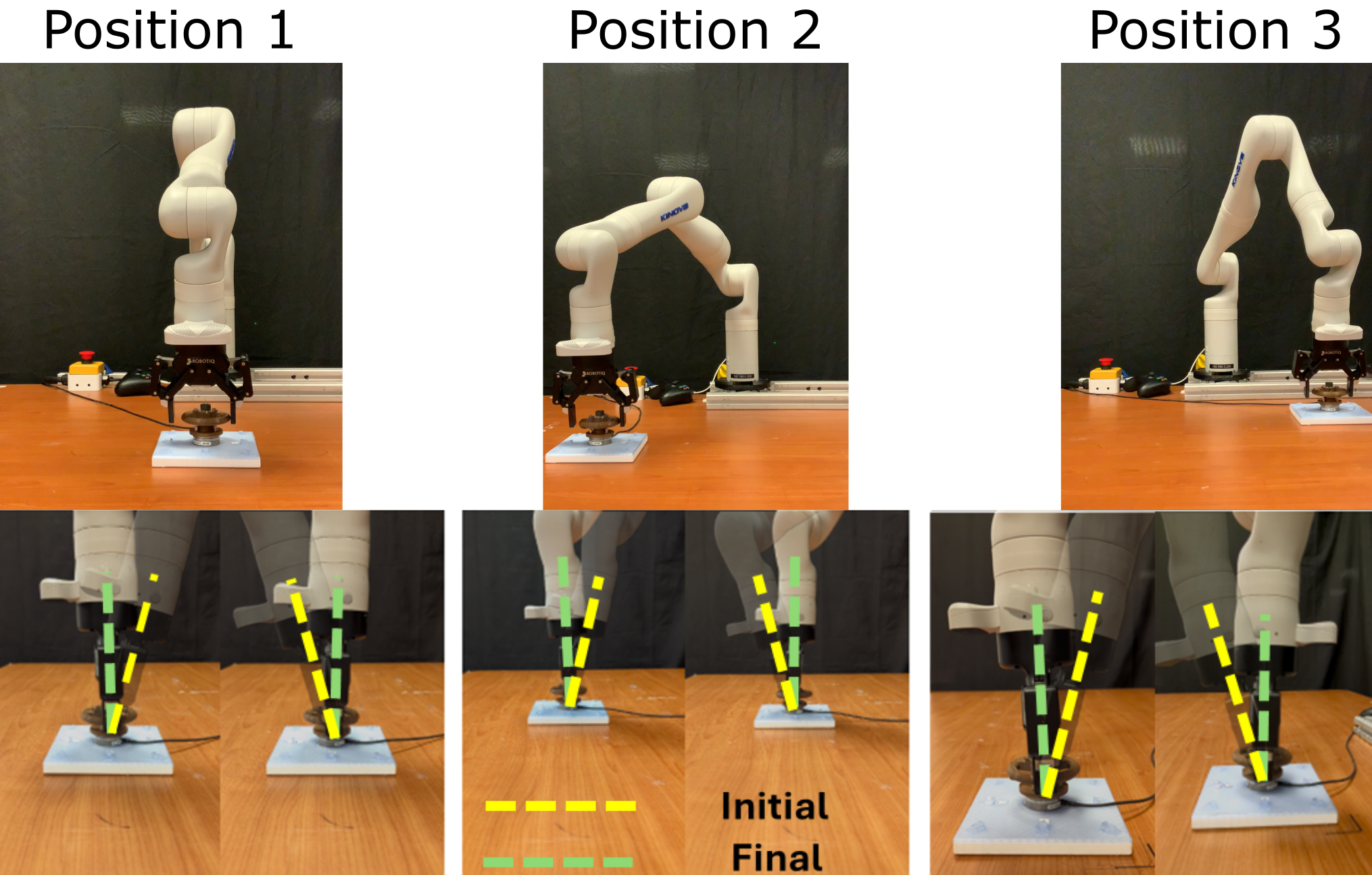}
    \caption{(Top) Different positions of the valve. (Bottom) Final orientation of end-effector (green) for $\theta_{y_0}=\pm 15^\degree$(yellow).}
    \label{fig:exp_hw_demons}
\end{figure}

\begin{table}[tb]
\centering
\caption{Closed loop Drake simulation results for ten initial axial misalignment conditions.}
\label{tab:drake_batch}
\scriptsize 
\begin{tabular}{cccccc}
\toprule
Case & $n_{1,0}$ & $n_{2,0}$ & End $\|\boldsymbol{\tau}_{xy}\|$ (Nm) & End $\|\mathbf{n}\|$ & Time (s) \\
\midrule
1  &  0.0000 & -0.0524 & 0.00069 & 0.00272 & 22.3 \\
2  &  0.0000 &  0.0524 & 0.00078 & 0.00224 & 23.7 \\
3  &  0.0524 &  0.0000 & 0.00087 & 0.00299 & 23.0 \\
4  & -0.0524 &  0.0000 & 0.00090 & 0.00245 & 23.0 \\
5  &  0.0349 & -0.0524 & 0.00070 & 0.00235 & 24.0 \\
6  & -0.0349 & -0.0524 & 0.00075 & 0.00223 & 24.9 \\
7  &  0.0524 &  0.0349 & 0.00122 & 0.00280 & 23.5 \\
8  & -0.0524 &  0.0349 & 0.00061 & 0.00232 & 24.3 \\
9  &  0.0699 & -0.0262 & 0.00077 & 0.00261 & 24.5 \\
10 & -0.0262 &  0.0699 & 0.00084 & 0.00281 & 24.1 \\
\midrule
Mean & -- & -- & 0.00081 & 0.00255 & 23.7 \\
\bottomrule
\end{tabular}
\end{table}

To further test the robustness of the feedback behavior, ten different initial misalignment conditions were simulated. These cases include both single-axis and combined-axis perturbations in $n_1$ and $n_2$. The results are summarized in Table \ref{tab:drake_batch}. All ten cases reached the torque plane convergence threshold. The convergence time remained within a narrow range of approximately 22-25 seconds, despite the different initial directions and magnitudes of misalignment. The final displayed torque plane distance was below $1.3\times10^{-3}\,\mathrm{Nm}$ in all cases, indicating that the last torque ellipse had collapsed close to the origin. The final values of $\|\V n\|$ were also small, showing that the feedback update consistently reduced the axial pose error.

\section{Hardware Experiment}
\label{sec:hw_exp}
This experiment was performed using a Robotiq 2F-85 gripper attached to a 7-DoF Kinova Gen3 robotic arm as shown in Figure~\ref{fig:exp_hw_demons}. The important frames of reference are shown in Figure~\ref{fig:valve_model_v2}. The transformation between the end-effector and the robot, $\V T_E^R$, is found using forward kinematics (FK). Since the end-effector and the fingers of the gripper are separated by a distance, we post-multiply $\V T_G^E$ with $\V T_E^R$ to get the transformation from $\{G\}$ to $\{R\}$. Since $\V T_V^R$ is fixed, we get $\V T_V^G$:
\begin{align}
    \V T_E^R &= \text{FK}(q) &: \{E\} \to \{R\}\nonumber \\
    \V T_G^R &= \V T_E^R \V T_G^E &: \{G\} \to \{R\} \nonumber \\
    \V T_G^V &= ((\V T_G^R)^{-1}\V T_V^R)^{-1} &: \{G\} \to \{V\}   
    \label{eq:T_V_to_G}
\end{align}
$\V T_G^V$ consists of the rotation matrix, $\V R_G^V$, and the displacement vector, $\V p_G^V$. The displacement remains the same during correction of the axial misalignment. The third column of $\V R_G^V$ is divided by the third component of the same column along with its norm to get $\V z_G$ in the form of eq.~\ref{eq:zg_def}.

\begin{figure}[tb]
    \centering
    \includegraphics[width=\linewidth]{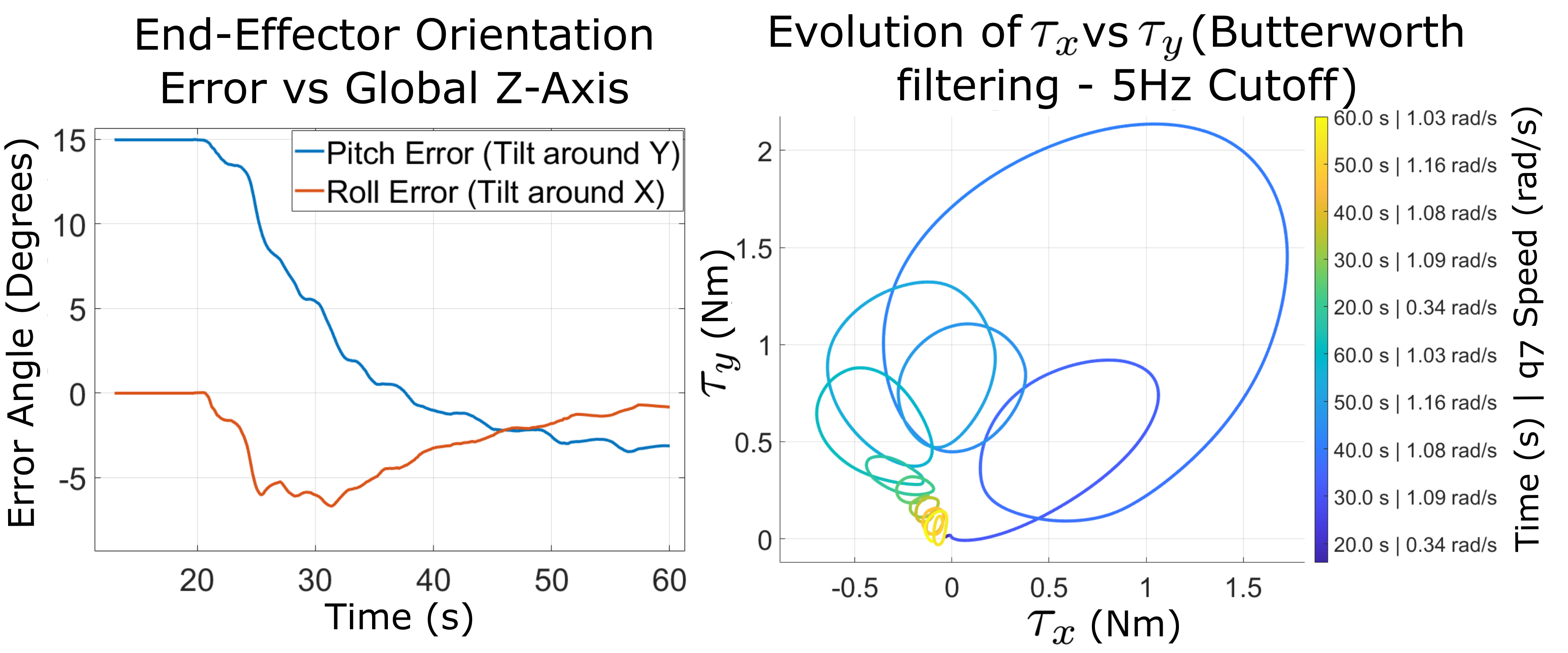}
    \caption{Evolution of (Left) roll and pitch error, (Right) filtered $\tau_x$ vs $\tau_y$ in the hardware setup for initial misalignment of 15 degrees. The color bar on the right represents time and the end-effector's speed.}
    \label{fig:axial_tau_error_hw}
\end{figure}

An ATI Mini40 F/T sensor is mounted beneath the valve. The $z$-component of the measured torque from the ATI F/T sensor is removed to capture only the reaction torques experienced by the gripper.
\begin{align}
  \boldsymbol{\tau_{G}}^V = 
  \begin{bmatrix} 1 & 0 & 0\\ 0 & 1 & 0\\ 0 & 0 & 0 \end{bmatrix}
  \begin{bmatrix} \tau_x \\ \tau_y \\ \tau_z \end{bmatrix}
  \label{eq:tauG_in_V}
\end{align}

\begin{figure}[tb]
    \centering
    \includegraphics[width=0.85\linewidth]{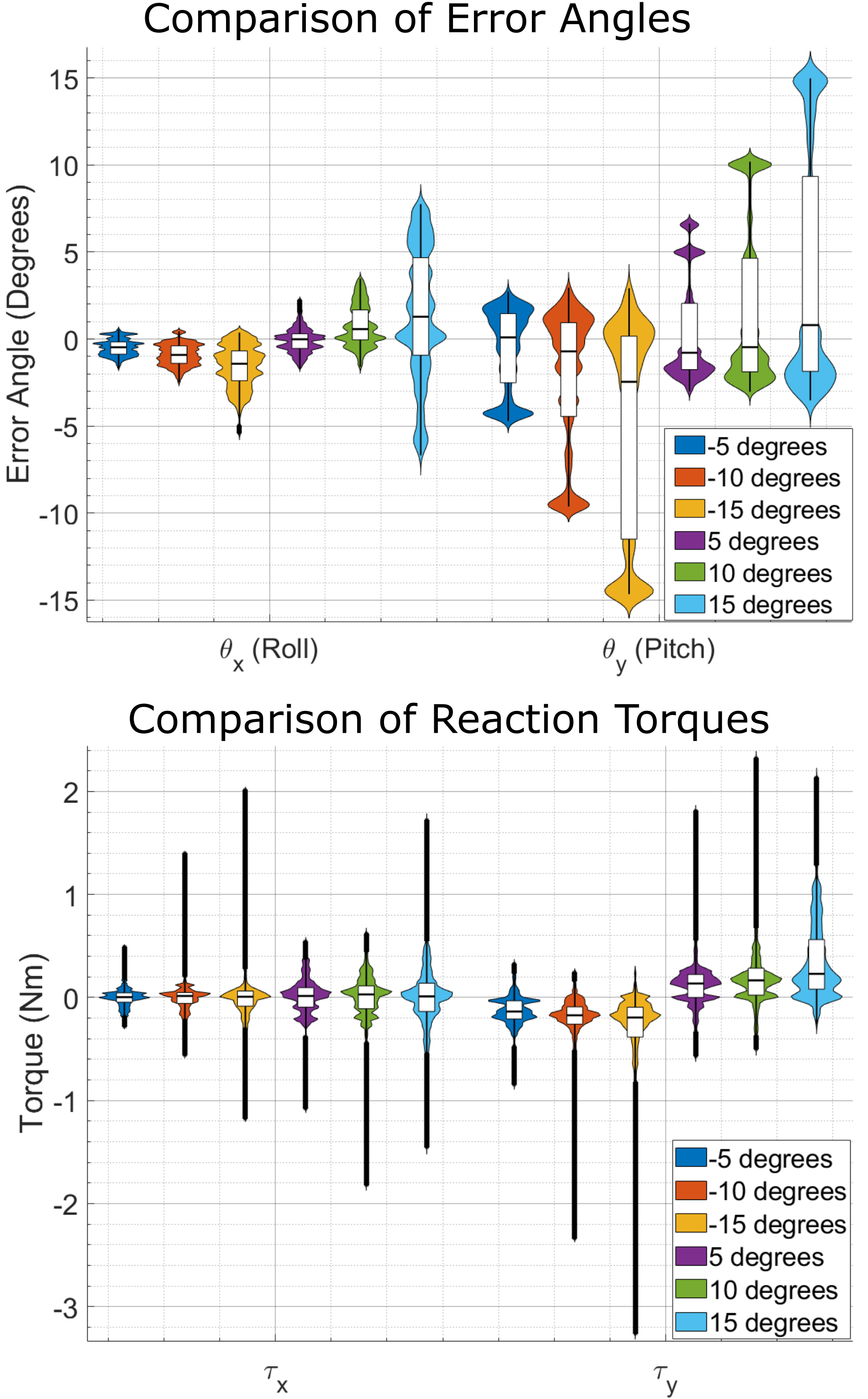}
    \caption{Distribution of errors and reaction torques for 18 test cases with $\theta_{y_0}=[- 15^\degree, 15^\degree]$ observed across three different valve positions in the hardware experiment. (Top)~Distribution of $\theta_x$ and $\theta_y$. (Bottom) Distribution of $\tau_x$ and $\tau_y$.}
    \label{fig:all_five_no_summ_results}
\end{figure}

Using~\cref{eq:z_G_dot,eq:y_G_dot,eq:x_G_dot,eq:R_G_dot,eq:Omega,eq:omega_in_R,eq:q_t_update}, the next $\mathbf{q}$ is fed back to Kinova in position control mode using a stiffness coefficient of 400~$\mathrm{Nm}(\mathrm{rad})^{-1}$ and a frequency of 100~Hz. $\beta$ is set to 0.18~$\mathrm{rad(Nm} \cdot \mathrm{s})^{-1}$. The evolution of the pitch ($\theta_y$) and the roll ($\theta_x$) error with respect to the z-axis and $\tau_x$ vs $\tau_y$, for an initial misalignment ($\theta_{y_0}$) of $15\degree$, is shown in Figure~\ref{fig:axial_tau_error_hw}. Note that Butterworth filtering is used only for plotting. The error in $\theta_y$ reduces substantially from $14.98\degree$ to $3.11\degree$ along with the size of the double loops. Each misalignment is repeated across three different valve positions. Each experiment lasted roughly 47 seconds. An example of two such test cases, $\theta_{y_0} \text{ as } -15\degree$ and $\theta_{y_0} \text{ as } 15\degree$, tested at 3 different valve positions, is shown in Figure~\ref{fig:exp_hw_demons}. Figure~\ref{fig:all_five_no_summ_results} shows the distribution of these parameters for different $\theta_{y_0}$ ranging from $-15\degree$ to $+15\degree$. As $\theta_{y_0}$ is increased, the maximum or the minimum $\tau_x$ and $\tau_y$ (depending on the direction) experienced by the end-effector, increases proportionally. Median of the distribution of $\theta_y$ remains within $[-2.46, 0.80]\degree$, $\theta_x$ within $[-1.42, 1.28]\degree$, $\tau_x$ within $[0.01, 0.03]$Nm, and $\tau_y$ within $[-0.19, 0.23]$Nm. These discrepancies may stem from inaccuracies in the manipulator's forward kinematic model and the mechanical wear of the 3D-printed valve.

\section{CONCLUSIONS}

In this work, we present a haptic update control law for the axial alignment of a robotic gripper and a valve, based on the geometric relationship between the reaction torque ellipses and the misalignment vector. We proved that our system dynamics are persistently exciting and are uniformly asymptotically stable. The results highlight the simplicity and robustness of using geometric relationships to construct control laws for contact-rich tasks. Future work will focus on making the rotational admittance a function of the state and testing such control laws on soft continuum robots.

\addtolength{\textheight}{-1cm}   



\appendix

\subsection{Haptic Update Control Law in terms of $\dot{\V n}$}
\label{sec:app_haptic_up_rule_n_dot}

We re-write~\cref{eq:z_G_dot} in terms of the free variables $\V n = \MAT{n_1\\n_2}$:
\begin{equation}
    \dot{\V z}_G = J \dot{\V n} = - \beta \cdot \V \tau_G(n_1,n_2,\theta) \times \V z_G.
    \label{eq:app_n1n2_dot}
\end{equation}
where, the $3\times2$ Jacobian $J = \nabla_{\V n} \V z_G$ is given by:
\begin{equation}
J = \MAT{\frac{1}{\sqrt{n_1^2+n_2^2+1}}-\frac{n_1^2}{(n_1^2+n_2^2+1)^{3/2}} & \frac{-n_1n_2}{(n_1^2+n_2^2+1)^{3/2}} \\
\frac{-n_1n_2}{(n_1^2+n_2^2+1)^{3/2}} & \frac{1}{\sqrt{n_1^2+n_2^2+1}}-\frac{n_2^2}{(n_1^2+n_2^2+1)^{3/2}}\\
\frac{-n_1}{(n_1^2+n_2^2+1)^{3/2}} & \frac{-n_2}{(n_1^2+n_2^2+1)^{3/2}}}
\label{eq:app_Jacobian_n1n2_z_G}
\end{equation}
To simplify further, we introduce $\sigma = \sqrt{n_1^2 + n_2^2 + 1}$ and multiply it on both sides of eq. \ref{eq:app_n1n2_dot}:
\begin{equation}
    J_\sigma \dot{\V n} = - \sigma \beta \cdot \V \tau_G(n_1,n_2,\theta) \times \V z_G.
    \label{eq:app_n1n2_dot_with_sigma}
\end{equation}
where, $J_\sigma = \sigma J$ is:
\begin{equation}
    J_\sigma = \frac{1}{\sigma^2}\MAT{n_2^2 + 1 & -n_1n_2\\
    -n_1n_2 & n_1^2 + 1\\
    -n_1 & -n_2
    }
    \label{eq:app_J_sigma}
\end{equation}
Since we have two free variables, we extract the $2\times2$ minor of $J_\sigma$ matrix in eq.~\ref{eq:app_J_sigma} as:
\begin{equation}
    J_m = \Pi_{x,y}J_\sigma
    \label{eq:app_J_m}
\end{equation}
where, $\Pi_{x,y} = \MAT{1 & 0 & 0\\ 0 & 1 & 0}$. Eq.~\ref{eq:app_n1n2_dot_with_sigma} is then re-written by multiplying $\Pi_{x,y}$ on both sides:
\begin{equation}
    J_m \dot{\V n} = - \sigma \beta \cdot \Pi_{x,y} \cdot \V \tau_G(n_1,n_2,\theta) \times \V z_G.
    \label{eq:app_n1n2_dot_with_J_m}
\end{equation}
and finally using $()^\wedge$ operator:
\begin{equation}
\dot{\V n} = \sigma \beta \cdot J_m^{-1} \cdot \Pi_{x,y} \cdot \widehat{\V z}_G \cdot \tau_G(n_1,n_2,\theta).
\label{eq:app_Jacobian_n1n2_z_G_simplified}
\end{equation}
where, the inverse of $J_m$ is:
\begin{equation}
    J_m^{-1} = \MAT{n_1^2+1 & n_1n_2\\
    n_1n_2 & n_2^2 +1}
    \label{eq:app_J_m_inv}
\end{equation}
Using~\cref{eq:tau_G_n1_n2_theta}, we get the components of $\V \tau_G$ as:
\begin{align}
\tau_x &= \frac{1}{\rho_1 \rho_2 \sqrt{\rho_3}} \Big( 2k n_1^2 n_2 \mathrm{r_G} \mathrm{r_V} \cos^2(\theta) - 2k n_2 \mathrm{r_G} \mathrm{r_V} \sin^2(\theta) \rho_3 \nonumber \\
       &\quad - k n_1 \mathrm{r_G} \mathrm{r_V} \sin(2\theta) \sqrt{\rho_3} + k n_1 n_2^2 \mathrm{r_G} \mathrm{r_V} \sin(2\theta) \sqrt{\rho_3} \Big) \label{eq:app_tau_x_alpha_star} \\
\tau_y &= \frac{2k \mathrm{r_G} \mathrm{r_V} \cos(\theta) \big(n_1 \cos(\theta) + n_2 \sin(\theta) \sqrt{\rho_3}\big) \rho_1}{\rho_2 \sqrt{\rho_3}} \label{eq:app_tau_y_alpha_star}
\end{align}
where,
\begin{align*}
\rho_1 &= \sqrt{n_2^2 + 1} \\
\rho_2 &= \sqrt{\cos^2(\theta) (n_2^2 + 1)^2 + \big(\sin(\theta) \sqrt{\rho_3} - n_1 n_2 \cos(\theta)\big)^2} \\
\rho_3 &= n_1^2 + n_2^2 + 1
\end{align*}
Substituting $\V \tau_G$ in~\cref{eq:app_Jacobian_n1n2_z_G_simplified}, we get the components of $\dot{\V n}$ as:
\begin{align}
\dot{n}_1 &= -\frac{\beta kn_1 n_2 }{\rho_1 \rho_2} \Big( n_1 \mathrm{r_G} \mathrm{r_V} \sin(2\theta) \sqrt{\rho_3} - 2n_1^2 n_2 \mathrm{r_G} \mathrm{r_V} \cos^2(\theta) \nonumber \\
    &\quad + 2n_2 \mathrm{r_G} \mathrm{r_V} \sin^2(\theta) \rho_3 - n_1 n_2^2 \mathrm{r_G} \mathrm{r_V} \sin(2\theta) \sqrt{\rho_3} \Big)- \nonumber \\
    &\quad \frac{2\beta k \mathrm{r_G} \mathrm{r_V} \cos(\theta) \big(n_1 \cos(\theta) + n_2 \sin(\theta) \sqrt{\rho_3}\big) (n_1^2 + 1) \rho_1}{\rho_2} \nonumber \\
    &= \frac{2\beta k \mathrm{r_G} \mathrm{r_V} \rho_4}{\rho_1 \rho_2}\Big(n_1 n_2(n_1 n_2\cos(\theta)-\sqrt{\rho_3}\sin(\theta)) \nonumber \\ 
    &\quad -\cos(\theta)(n_1^2+1)\rho_1^2 \Big)     \label{eq:app_n1_dot} \\[2ex]    
    \dot{n}_2 &= -\frac{2\beta k \mathrm{r_G} \mathrm{r_V} (n_2  \sin^2(\theta) \rho_1 \rho_3 + n_1 \cos(\theta) \sin(\theta) \rho_1 \sqrt{\rho_3})}{\rho_2} \nonumber \\
    &= -\frac{2\beta k \mathrm{r_G} \mathrm{r_V} \rho_4 \rho_1 \sqrt{\rho_3}\sin(\theta)}{\rho_2}
    \label{eq:app_n2_dot}   
\end{align}
where,
\begin{align*}
\rho_4 &= n_1\cos(\theta)+n_2\sin(\theta)\sqrt{\rho_3}
\end{align*}

\subsection{Derivative of Lyapunov function}
\label{sec:deri_lyapunov_func}
Substituting~\cref{eq:app_n1_dot,eq:app_n2_dot} in~\cref{eq:dot_v}, we calculate $\dot{V}$:

\begin{align}
    \dot{V}(\V n, t) &= \frac{n_12\beta k \mathrm{r_G} \mathrm{r_V} \rho_4}{\rho_1 \rho_2}\Big(n_1 n_2(n_1 n_2\cos(\theta)-\sqrt{\rho_3}\sin(\theta)) \nonumber \\
    & \quad -\cos(\theta)(n_1^2+1)\rho_1^2 \Big) \nonumber \\ 
    & \quad -\frac{n_22\beta k \mathrm{r_G} \mathrm{r_V} \rho_4 \rho_1 \sqrt{\rho_3}\sin(\theta)}{\rho_2} \nonumber \\
    & = -\frac{2\beta k \mathrm{r_G} \mathrm{r_V} \rho_3 \rho_4^2}{\rho_1 \rho_2} \nonumber \\
    & = - \Gamma(\V n, t)\cdot(\mathbf{w}^{\top}(\V n, t)\V n)^2
    \label{eq:subs_dot_v}
\end{align}
where,
\begin{align*}
    \Gamma(\V n, t) &= \frac{2\beta k r_G r_V \rho_3}{\rho_1 \rho_2}, \\
     \mathbf{w}(\V n, t) &= [\cos(\omega^G t) \ \sin(\omega^G t)\sqrt{\rho_3}]^{\top}
\end{align*}


\bibliographystyle{IEEEtran}
\bibliography{mybibfile}

\end{document}